%% file: main.tex
\documentclass[10pt,twocolumn,letterpaper]{article}

\usepackage{cvpr}              


\definecolor{cvprblue}{rgb}{0.21,0.49,0.74}
\usepackage[pagebackref,breaklinks,colorlinks,allcolors=cvprblue]{hyperref}
\usepackage[accsupp]{axessibility}
\usepackage{xcolor}         
\usepackage{multirow}
\usepackage{amsmath}

\def\paperID{16943} 
\def\confName{CVPR}
\def\confYear{2025}

\title{RANGE: Retrieval Augmented Neural Fields for Multi-Resolution Geo-Embeddings}

\author{Aayush Dhakal$^1$
Srikumar Sastry$^1$
Subash Khanal$^1$
Adeel Ahmad$^{1,2}$
Eric Xing$^1$
Nathan Jacobs$^1$\\[\bigskipamount]
Washington University in St. Louis$^1$ \quad Taylor Geospatial Institute$^2$\\
 }

\begin{document}
\maketitle
\input{sec/0_abstract}

\input{sec/1_intro}

\input{sec/related_work}
\input{sec/3_method}
\input{sec/4_results}
\input{sec/5_conclusion}

{
    \small
    \bibliographystyle{ieeenat_fullname}
    \bibliography{main}
}

\input{sec/X_suppl}


\end{document}

%% file: sec/0_abstract.tex
\begin{abstract}

The choice of representation for geographic location significantly impacts the accuracy of models for a broad range of geospatial tasks, including fine-grained species classification, population density estimation, and biome classification. Recent works like SatCLIP and GeoCLIP learn such representations by contrastively aligning geolocation with co-located images. While these methods work exceptionally well, in this paper, we posit that the current training strategies fail to fully capture the important visual features. We provide an information theoretic perspective on why the resulting embeddings from these methods discard crucial visual information that is important for many downstream tasks. To solve this problem, we propose a novel retrieval-augmented strategy called RANGE. We build our method on the intuition that the visual features of a location can be estimated by combining the visual features from multiple similar-looking locations. We evaluate our method across a wide variety of tasks. Our results show that RANGE outperforms the existing state-of-the-art models with significant margins in most tasks. We show gains of up to 13.1\% on classification tasks and 0.145 $R^2$ on regression tasks. All our code and models will be made available at: \href{https://github.com/mvrl/RANGE}{https://github.com/mvrl/RANGE}.

\end{abstract}

%% file: sec/1_intro.tex
\section{Introduction}
\label{sec:intro}

\begin{figure}[!t]
\begin{center}
\includegraphics[width=\linewidth, scale=0.3]{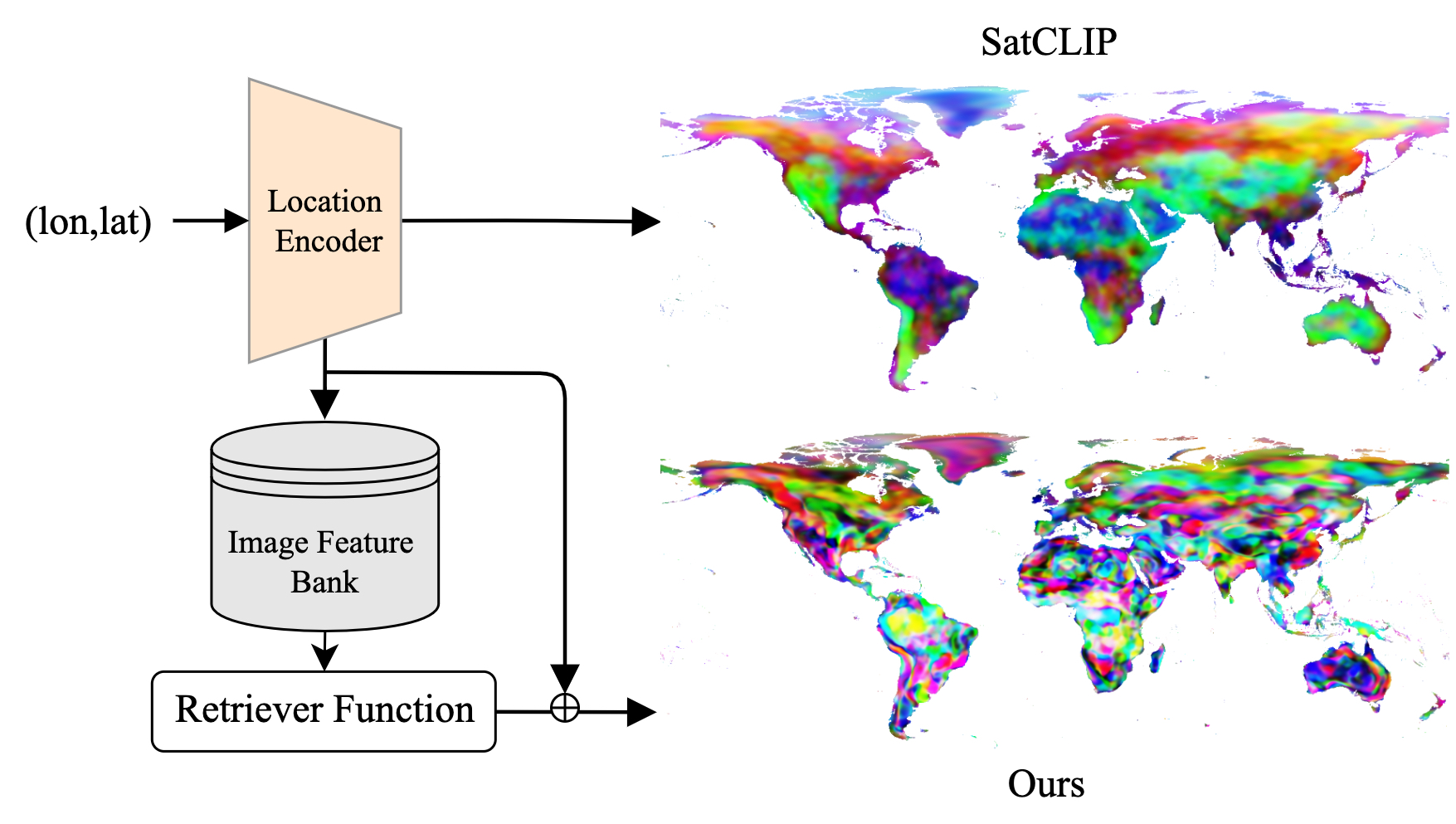}
\end{center}
   \caption{Adding explicit visual features allows us to generate more high-resolution location embeddings. We use a retrieval strategy that allows us to generate multi-resolution retrieval-augmented neural field of geo-embeddings (RANGE).} 
\label{fig:multiview}
\end{figure}

Several machine learning tasks require the use of geolocation as an input feature. Multiple works have shown the benefits of using geolocation to solve vital ecological, geographical, and geological tasks. SINR~\cite{cole2023spatial} used location as input to predict species distribution, Lang \textit{et al.}~\cite{lang2023high} used location with satellite image to learn a global canopy height model, and SatCLIP~\cite{klemmer2023satclip} used location to solve several geospatial tasks such as biome classification, population density estimation, housing price prediction, etc. Since location is a ubiquitous input variable in all geospatial settings, finding good representations for it is fundamental when solving geospatial tasks. 

Previous work~\cite{russwurm2023geographic} has shown that using location directly (represented as two floating point numbers of latitude and longitude) as an input to a machine learning model yields poor performance. Therefore, several works have attempted to find better ways to represent location data. Some of these approaches~\cite{Aodha_2019_ICCV,gaolearning} use a non-parametric method to encode location. However, recent works~\cite{vivanco2024geoclip,mai2023csp,klemmer2023satclip} have shown that it is possible to learn a general-purpose location representation by learning a parametric model (often a neural network) in a self-supervised setting. There are several strategies that can be used to train the parametric model. A common approach used by state-of-the-art models~\cite{vivanco2024geoclip,mai2023csp,klemmer2023satclip} involves learning visual cues associated with a given geolocation. Two recent state-of-the-art models, GeoCLIP~\cite{vivanco2024geoclip} and SatCLIP~\cite{klemmer2023satclip}, use contrastive training to align geo-embeddings with co-located images; the former aligns them with ground-level images while the latter aligns them with satellite images. Images of a location capture dense information about the area, such as the landcover, landuse, etc. Therefore, learning location representations that capture this dense information is beneficial for solving geospatial tasks. Prior work~\cite{klemmer2023satclip} has shown that representations from such models are extremely powerful and are capable of solving diverse geospatial tasks both on their own and in combination with images. In this work, we argue that the current approaches to learning geo-embeddings using images, while powerful, still fail to preserve all the relevant visual information. The location representations from these models only capture low-resolution shared information between location and image and ignore crucial high-resolution information that is unique to the image. In section~\ref{sec:multiview}, we define this problem more formally from the perspective of multi-view redundancy. 

To address this issue, we propose a retrieval-augmented strategy that approximates the unique visual information present in a location. There are three key ideas to our solution. First, it is possible to approximate the visual features of a given location by strategically combining the visual features of other similar-looking locations. Second, contrastively trained models like SatCLIP~\cite{klemmer2023satclip} and GeoCLIP~\cite{vivanco2024geoclip}, while lacking in preserving modality-specific information, are excellent in learning modality alignment~\cite{gupta2022understanding}. Therefore, it is possible to retrieve semantically aligned images, with alignment scores, for a given location using these models. Third, we note that the variance of semantic information visible in satellite images across the Earth is relatively low (as opposed to the wider diversity in consumer photographs). Hence, it is possible to capture a large percentage of unique aerial semantics with a limited number of satellite images. By combining these three key ideas, we propose our retrieval-augmented framework, RANGE (Retrieval-Augmented Neural Fields for Multi-Resolution Geo-Embeddings). Our retriever function uses both semantic and spatial alignment to approximate the visual features of a given location using an auxiliary database of image features. Our function is robust to the database size and works well even in limited sample settings. The hyperparameter in the retriever function can be adjusted to generate geo-embeddings at different frequencies. In summary, these are the main contributions of our paper:   

\begin{itemize}
\item We show why existing image-location contrastive methods result in sub-optimal representations from the perspective of multi-view redundancy.  
\item We propose a novel retrieval-based strategy, RANGE, to generate geo-embeddings that preserve both shared and unique modality information.
\item We propose a retriever function that estimates the visual embedding of a location using both semantic and spatial alignment. Our retriever function is robust to different database sizes and can generate geo-embeddings at multiple frequencies by enforcing spatial smoothness.   
\end{itemize}

\section{Aligning Location and Image is a Multi-View Non-Redundancy Problem}
\label{sec:multiview}
Let (L,H,T) represent location modality, image modality, and geospatial task, respectively. Models like SatCLIP and GeoCLIP align co-located images and locations by minimizing the infoNCE objective. Solving the multimodal infoNCE objective maximizes the lower bound on the mutual information, preserving the shared information within the learned representations~\cite{oord2018representation}. As such, the final learned representations of these models contain only information that is shared across both location and image. This property is useful under the multi-view redundancy assumption which states that the information shared between the two modalities is exactly relevant to solve the underlying task.
In other words, the multi-view redundancy setting assumes that any additional unique information from the image or location would not be useful in solving the task T~\cite{tosh2021contrastive}. However, this assumption does not hold in the location-image learning setting as there is high unique information contained in the image modality that is useful for solving downstream tasks. Formally, we argue that location-image learning falls under the multi-view non-redundancy setting: there is \textit{unique} information in the image modality that is not shared with the location modality, and that information is relevant for solving our geospatial task T. In this case, contrastive training discards valuable task-relevant information from the image, leading to poor downstream performance~\cite{liang2023factorized}.

Empirically, there is abundant unique information present in the image features that are relevant for solving downstream tasks. The location representations from contrastively trained models like SatCLIP and GeoCLIP only capture the low-resolution shared information, ignoring all the valuable task-relevant information that is unique to the image. To demonstrate this property concretely, we quantitatively show that there are tasks that can be solved more accurately by adding the image representation to the shared image-location representation from SatCLIP~\cite{klemmer2023satclip}. First, we choose 4 tasks: Biome, Ecoregions, Elevation, and Population Density estimation. We download a corresponding Sentinel-2 image for every data point. For each task, we train one model using the SatCLIP features and another model using the SatCLIP features combined with the SatMAE~\cite{cong2022satmae} features of the satellite image~\cite{cong2022satmae}. The results in Table~\ref{table:multiview} show that adding image features improves the performance across all tasks. The increase in performance demonstrates that the images contain unique information relevant to solving the task that is lost during contrastive training. Hence, ``location and image" does not follow the assumption of multi-view redundancy.

Although adding corresponding satellite image information to a location embedding improves performance, this is not feasible in practice. For many geospatial tasks, we wish to make predictions across the globe with millions of points. In such cases, using the ``image+location" framework would require us to retrieve/store/process millions of images. We propose an efficient method for approximating the visual information using a compact database. Our results suggest that this approximation can outperform the use of true visual features in some cases.

\begin{table}[]
\begin{tabular}{l|cc|cc}
             & Biome & Ecoregion & Elevation & Population \\ \hline
SatMAE  & 58.8                      & 28.4                           & 0.388 & 0.600 \\
SatCLIP     & 68.9  & 69.3      & 0.666     & 0.684      \\
loc$\oplus$img & 74.9  & 73.5      & 0.749     & 0.765      \\ \hline
gain (\%)             & +8.71   & +6.06       & +12.46     & +11.84  
\end{tabular}
\caption{We show that using the image-location shared information from SatCLIP embeddings is sub-optimal to solve some geospatial problems. Adding image features to the embeddings provides useful visual information that improves the accuracy of the task. We show the accuracy for the classification tasks and R$^2$ value for the regression tasks. The results indicate that there are valuable visual features that are not captured by the SatCLIP embeddings.}
\label{table:multiview}
\end{table}

%% file: sec/related_work.tex
\section{Related Works}
\subsection{Learning Representations for Geolocation}
Learning effective representations for geolocation is valuable for diverse downstream geospatial tasks, such as geolocalization~\cite{vivanco2024geoclip}, canopy height prediction~\cite{lang2023high}, fine-grained species classification~\cite{Aodha_2019_ICCV,sastry2024birdsat}, species distribution modeling~\cite{cole2023spatial, sastry2024birdsat}, geographical question answering \cite{mai2020se}, soundscape mapping~\cite{khanal2024psm}, etc. Geolocation in these tasks can be represented using either non-parametric~\cite{gaolearning, russwurm2023geographic} or parametric~\cite{mai2023csp,vivanco2024geoclip,klemmer2023satclip} methods.
Ru{\ss}wurm \textit{et al.}~\cite{russwurm2023geographic} represents geolocation using spherical harmonic basis functions, and Aodha \textit{et al.}~\cite{Aodha_2019_ICCV} uses a sine and cosine encoding of geolocation that removes discontinuity at dateline. In recent years, learned representations for geolocation, typically parameterized by neural networks, have been popularized. Some of these methods learn geolocation representations by using self-supervised pre-training strategies that distill visual information from either co-located overhead~\cite{klemmer2023satclip,mai2023csp} or ground-level images~\cite{mai2023csp,vivanco2024geoclip}. The location encoders trained by these methods have demonstrated utility across various downstream geospatial tasks. We use the representations from these location encoders as baselines for our model.
\subsection{Multimodal Contrastive Learning}
Contrastive learning is an efficient and scalable pre-training strategy for learning a shared embedding space across multiple modalities~\cite{radford2021learning,yang2022vision,yaofilip,li2021align,girdhar2023imagebind,elizalde2023clap,Dhakal_2024_CVPR,khanal2024psm,huynh2024contrastive,zheng2023exif}. As demonstrated by CLIP~\cite{radford2021learning} and more recent works~\cite{li2021align,yaofilip,yang2022vision} in vision-language learning, the shared embedding space can be utilized for zero-shot classification and leveraged across a variety of downstream tasks. Contrastively trained multimodal embedding spaces between geolocation and satellite imagery~\cite{klemmer2023satclip, mai2023csp} or ground-level imagery~\cite{vivanco2024geoclip}, have also demonstrated impressive transferability of both location and visual representations to various geospatial tasks.

Despite its success, recent works highlight the limitations of multimodal contrastive learning. One such study~\cite{liang2023factorized} shows that contrastively trained representations capture the shared information between the two modalities and thus only work best under certain assumptions. Similarly, another work~\cite{gupta2022understanding} demonstrates that the final layers of encoders in the contrastive learning framework preserve only the information necessary for solving the alignment objective, discarding other modality-specific details. We argue that some of the underlying assumptions of multimodal contrastive training do not hold true in the image-location setting. Therefore, the resulting location representations from these methods are sub-optimal for downstream tasks. To address this issue, we propose a method to approximate the image-specific information for a given location.
\subsection{Retrieval Augmented Methods}
There has been a recent surge in retrieval-augmented generation (RAG) methods, primarily focused on generative tasks~\cite{chenre, blattmann2022retrieval, kirstain2023x, lewis2020retrieval, borgeaud2022improving, siriwardhana2023improving, seoretrieval, koizumi2020audio, wangretrieval}. These methods generally consist of three main components: a retriever, an external database, and a generator. For a given query, the retriever selects a set of documents from the database and provides them, along with the original query, to the generator to produce the desired output. RAG has been successfully employed to improve text generation for large language models (LLMs)~\cite{lewis2020retrieval, borgeaud2022improving, siriwardhana2023improving}, text-to-image generation~\cite{chenre, blattmann2022retrieval, kirstain2023x}, text-to-3D generation~\cite{seoretrieval}, audio captioning~\cite{koizumi2020audio}, and more. Within the RAG framework, one could also replace the generator with another task-specific component, such as a classifier~\cite{long2022retrieval}, to enhance performance by leveraging the rich information from the retrieved documents. Inspired by these retrieval-augmented methods, we design our framework, RANGE, which retrieves rich visual information from our database for a given geolocation query and utilizes it for various downstream geospatial tasks.

%% file: sec/3_method.tex
\begin{figure*}[!t]
\begin{center}
\includegraphics[width=\linewidth, scale=0.3]{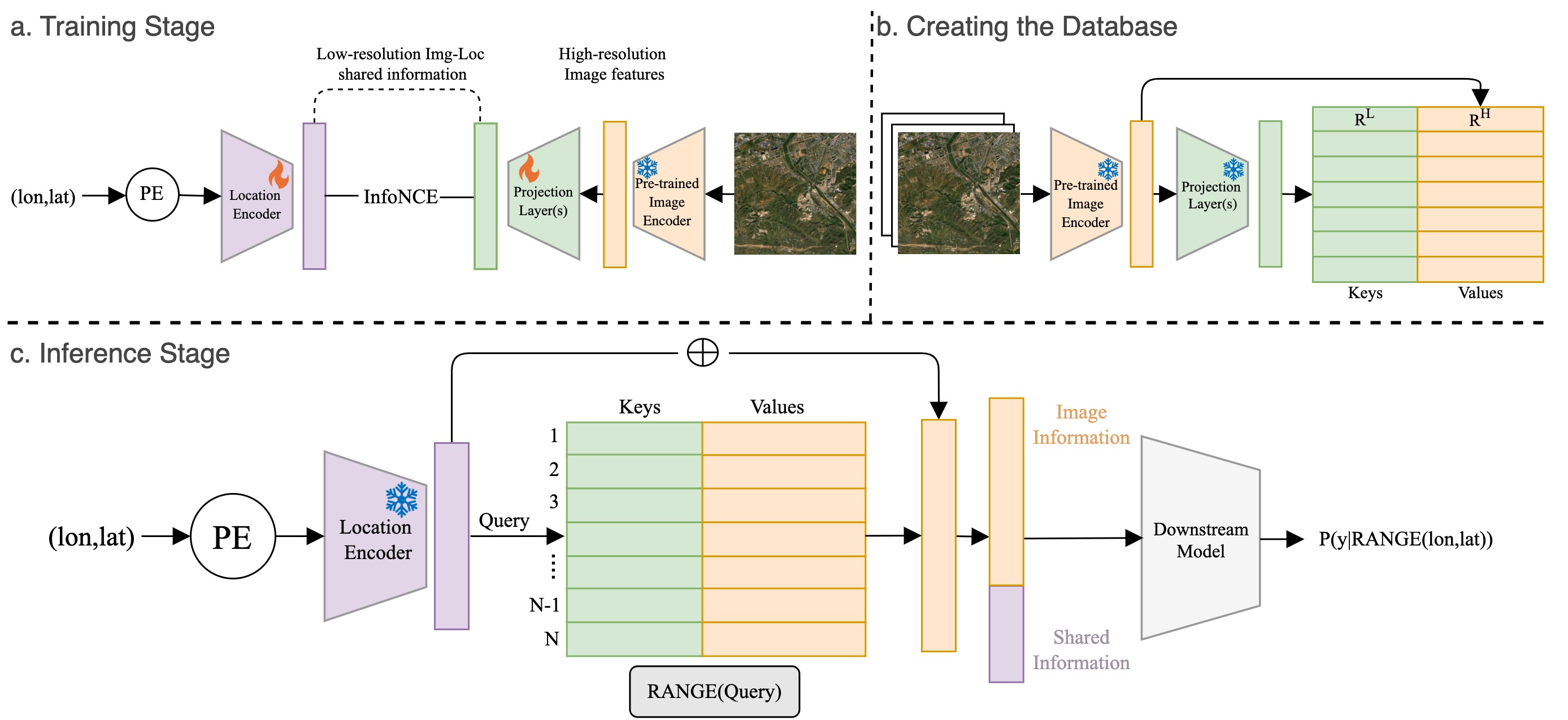}
\end{center}
    \caption{Framework of RANGE. (a) In the training stage, a shared embedding space is learned between locations and images. (b) We create a database of low-resolution and high-resolution image embeddings using the trained projection layer and a powerful pretrained image model, respectively. (c) During inference, we use a location as the query, low-resolution image embeddings as keys, and high-resolution image embeddings as values. Using our retriever function, we compute the approximate high-resolution embeddings for the query. We concatenate ($\oplus$) the approximated visual feature with our query embedding.} 
\label{fig:framework}
\end{figure*}
\section{Method}
\subsection{Problem Setup}
We consider a dataset with paired geolocations and co-located satellite images $\{g_i, s_i\}$. The goal is to learn a meaningful representation for $g_i$ by aligning it with $s_i$, which can be done by minimizing the clip objective \cite{klemmer2023satclip, vivanco2024geoclip, radford2021learning}. We have a trainable geolocation encoder $E$, a frozen pretrained image encoder $I$, and a projection layer $P$. Let $G_i = E(g_i)$ represent the location embedding, $S_i = {I(s_i)}$ represent the frozen image embedding, and $P_i = P(I(s_i))=P(S_i)$ represent the image embedding obtained from the projection layer for the i$^{th}$ sample. We minimize the objective:
\begin{align}
&L_{i}^{loc} = \frac{-1}{k}\sum_{i=1}^{k} \log\frac{\exp(G_i \cdot P_i / \tau)}{\sum_{j=1}^{k} \exp(G_i \cdot P_j) / \tau) } \\
&L_{i}^{img} = \frac{-1}{k}\sum_{i=1}^{k} \log\frac{\exp(G_i \cdot P_i / \tau)}{\sum_{j=1}^{k} \exp(G_j \cdot P_i) / \tau) } \\
&L_i = (L_i^{loc} + L_i^{img})/2 
\end{align}
The training process is shown in Figure~\ref{fig:framework}-(a). We contrastively align the geolocations and satellite images by minimizing their CLIP objective. This formulation is identical to SatCLIP~\cite{klemmer2023satclip}, so we use the pretrained SatCLIP image and location encoders for simplicity. For query location $g_i$, $S_i$ encodes the high-resolution visual features since $I$ is a powerful pre-trained image encoder. Both $P_i$ and $G_i$ encode low-resolution shared information between $g_i$ and $s_i$. The resulting geo-embedding, $G_i$, is sub-optimal as it loses unique visual information crucial for many downstream tasks. To solve this, we propose a retriever function $R_\tau$ that allows us to approximate the high-resolution information $S_i$ given $g_i$. Let $\epsilon$ denote the deviation from the true visual embedding. We can then explicitly add the approximate visual information $\hat{S_i}$ to our location embedding $G_i$:
    \begin{align}
    &R_\tau(g_i) = \hat{S_i} = S_i + \epsilon \\
    &RANGE(g_i) = \hat{S_i} \oplus G_i 
    \end{align}
\subsection{Retrieval Augmented Neural Fields}
\noindent \textbf{Creating the Retrieval Database: }Once the contrastive training is done, we create a retrieval database as shown in Figure~\ref{fig:framework}-(b). We take $\{g_i, s_i\}$ sampled uniformly across the globe using the SatCLIP~\cite{klemmer2023satclip} dataset. For every location $g_i$ in our data, we compute: a) the shared-information image embedding $R_i{^L}=P(I(s_i))$, and b) the high-resolution image embedding $R_i{^H}=\bar{I}(s_i))$; we refer to these sets of embeddings as low-resolution embeddings and high-resolution embeddings, respectively, in the rest of the paper. In practice, $\bar{I}$ can differ from $I$; the only condition is that $\bar{I}$ is a pretrained image feature extractor. For our purposes, we use SatMAE~\cite{cong2022satmae} as $\bar{I}$. 
\\
\noindent \textbf{Retriever Function: }
Let $R^L=\{R_{1}^L, R_{2}^L,...R_{N}^L\}$, and $R^H=\{R_{1}^H, R_{2}^H,...R_{N}^H\}$ represent the set of all low-resolution and high-resolution image embeddings in our database. After creating the database, we need a method to sample appropriate high-resolution image information for a query location. Naively, we can set this as a lookup operation, where $G_i$ is the query, $R^L$ is the set of all keys, and $R^H$ is the set of all values. We have a function $sim(G_i, R_{j}^{L})$, which gives the alignment scores between i and j. In our setting, this similarity function is a simple cosine similarity. For a query $G_i$, we find $R_{j}^{L}$ with the highest alignment score and return the key $R_{j}^{H}$. 
In practice, $R_{j}^{H}$ is noisy. The semantically closest image in the database can contain additional information that is irrelevant to the location. Therefore, naively adding $R_{j}^{H}$ has the potential of introducing incorrect information to our geo-embeddings.

We instead use a soft selection criteria. First, we compute the query's alignment score with each key in the database using cosine similarity. We use a softmax function to convert the alignment scores to probabilities. We also use a temperature parameter $\tau$ to slightly reshape the probability distribution; $\tau$ is extremely robust as it does not need to be fine-tuned for different tasks or for databases with different distributions. We use the resulting probabilities to compute a weighted average across all values, i.e., high-resolution image embeddings in the database. The resulting embedding is an approximation of $S_i$, where the information contributed by each image is weighted by its alignment score with the query. We concatenate this approximate high-resolution image embedding with our original location embedding $G_i$ to obtain the multi-resolution RANGE embedding $RANGE_i$. A high-level view of this process is shown in Figure~\ref{fig:framework}-(c). 

\begin{multline}
RANGE_i = R_{\tau}(G_i) \oplus G_i \\
= \frac{1}{N}\sum_{k=1}^{N}{\frac{e^{sim(G_i,R_{k}^L)/\tau}}
{\sum_{j=1}^{N}{e^{sim(G_i,R_{j}^L)/\tau}}}} * R_{k}^H \oplus G_i \\
\text{where}\ G_i\in \mathbb{R}^M,\ R_{k}^L \in \mathbb{R}^M,\ R_{k}^H \in \mathbb{R}^{N}.
\end{multline}
\noindent \textbf{Adding smoothness constraints: }We also propose another version of our model called $RANGE^+$. Here, we additionally use geodesic similarity to approximate the visual features of the location. We convert our query location to 3D cartesian coordinates and use angular distance to find the spatially closest image in the database. This spatial retrieval allows us to impose spatial smoothness on our embeddings since locations close to each other are forced to have similar retrieval, irrespective of the semantics.  Similar to $RANGE$, we compute a weighted average across all values in the database using the angular similarity between the query and keys. However, we explicitly mask samples with angular similarity lower than a threshold, i.e., significantly distant samples have no contribution. The spatially retrieved embedding is added to the semantically retrieved embedding and weighted using a $\beta$ parameter. This $\beta$ parameter controls the level of spatial smoothness enforced on the $RANGE$ embeddings, as shown in Figure~\ref{fig:beta}. Let $g{_i}^{3D}$ be the query location, $l$ be the set of all database locations in 3D cartesian coordinates, and $sim(g{_i}^{3D}, l_k)$ be their angular similarity:   
\begin{multline}
\label{eq:rangep}
RANGE^+_{i} = \frac{1}{N}(\beta*\sum_{k=1}^{N}{\frac{e^{sim(G_i,R_{k}^L)/\tau_{1}}}{\sum_{j=1}^{N}{e^{sim(G_i,R_{j}^L)/\tau_{1}}}}} * R_{k}^H+ \\
     (1-\beta)\sum_{k=1}^{N}{\frac{e^{sim(g{_i}^{3D},l_{k})/\tau_{2}}}{\sum_{j=1}^{N}{e^{sim(g{_i}^{3D},l_{j})/\tau_{2}}}}} * R_{k}^H
     )\oplus G_i \\
     \text{where } G_i \in \mathbb{R}^M, \ g{_i}^{3D} \in \mathbb{R}^3, \ l_k \in \mathbb{R}^3
\end{multline}
When $\beta$ is set to 0, only spatially relevant features are used, and we refer to this setting as RANGE-HAVER. 

%% file: sec/4_results.tex

\begin{table*}
\centering    

\begin{tabular}{l|ccc|ccccc}
\hline
                        & Biome & EcoRegions & Country   & Temperature & Elevation & Population & Cali-Housing   \\ \hline
Direct                      & 29.1 & 0.6 & 66.9& 0.381  & 0.025  &  0.053  & 0.238      \\
Cartesian\_3D                      & 30.2 & 1.8 &66.9 & 0.362  & 0.030  &  0.162  & 0.240      \\
Wrap~\cite{Aodha_2019_ICCV}                      & 34.4 & 1.1 & 69.7& 0.861  & 0.085  &  0.328  & 0.239      \\
Theory~\cite{gaolearning}                      & 33.5 & 1.0 &72.5 & 0.849  & 0.093  &  0.330  & 0.254      \\
SphereM~\cite{mai2023sphere2vec}                      & 36.4 & 27.3 & 72.7 & 0.629  & 0.139  &  0.302  & 0.423      \\
SphereM$^{+}$~\cite{mai2023sphere2vec}                      & 58.7 & 50.1 & 76.1 & 0.886  & 0.294  &  0.421  & 0.543      \\
SphereC~\cite{mai2023sphere2vec}                     & 36.3 & 52.9 & 72.9 & 0.461  & 0.185  &  0.335  & 0.496      \\
SphereC$^+$~\cite{mai2023sphere2vec}  & 53.2 & 61.6 & 73.6 & 0.842  & 0.260  &  0.392  & \underline{0.544}      \\

\hline

CSP-INat~\cite{mai2023csp}                      & 61.1 & 57.1 & 75.9 & 0.717  & 0.388  &  0.554  & 0.462      \\
CSP-FMoW~\cite{mai2023csp}                    & 61.4 & 58.0 & 81.3& 0.865  & 0.399  &  0.580  & 0.541      \\
SINR~\cite{cole2023spatial}                    & 67.9 & 54.9 & 88.3 & \textbf{0.942}  & 0.644  & 0.726 & 0.420      \\
GeoCLIP~\cite{vivanco2024geoclip}                & \underline{70.2} & 71.6 & 81.3 & 0.916  &  0.604  &  0.698  & \textbf{0.708}      \\
SatCLIP~\cite{klemmer2023satclip}  & 68.9 & 69.3 & 82.8 & 0.825       & 0.666     & 0.684 & 0.400   \\ \hline

RANGE  & \textbf{83.3} & \textbf{75.7} & \underline{93.7} &0.895 & \underline{0.844} & \underline{0.799} & 0.422  \\

RANGE$^+$                      & \textbf{83.3} & \underline{75.3} & \textbf{94.7} &\underline{0.931} & \textbf{0.851} & \textbf{0.811} & 0.336 &    \\ 
\end{tabular}
\caption{Our model shows improvements over a variety of tasks compared to state-of-the-art models. The first 3 columns are classification tasks with accuracy metrics and the last 4 columns are regression tasks with R$^2$ metric. We show improvements with significant margins across many of the tasks. Our model underperforms in the Cali-housing dataset. We further analyze this behavior in our discussions.}
\label{table:main}

\end{table*}

\begin{table*}
\resizebox{\linewidth}{!}{%
\begin{tabular}{l|cc|ccc|cccc}
\hline
                    &type &temp.   & Biome & Eco&Country  & Temperature & Elevation & Population & Cali-Housing   \\ \hline

SatCLIP & base-model &- & 68.9 & 69.3 & 82.8& 0.825       & 0.666     & 0.684 & 0.400   \\ 
\hline
RANGE & top-1&-     & 75.6 & 65.2 & 85.6 & 0.817      & 0.766    & 0.742 & 0.444  \\
&  top-k  &-& 82.8 & 76.8 & 90.6 & 0.884 & 0.810 & 0.771 & \textbf{0.619}  \\

 & soft selection & fixed& 83.3 & 75.7 &93.7 & 0.895 & 0.844 & 0.799 & 0.422  \\

& soft selection & per task& \underline{83.5} & \underline{75.8} & 94.5 & 0.922 & \textbf{0.857} & 0.809 & \underline{0.465}  \\ 

\hline

    RANGE$^+$ &       soft selection & fixed           & \underline{83.5} & 75.3 &\underline{94.7}  & \underline{0.931} & 0.851 & \underline{0.811} & 0.336    \\ 
    &       soft selection & per task           & \textbf{83.7} & \textbf{75.9} & \textbf{94.9} & \textbf{0.932} & \underline{0.855} & \textbf{0.813} & 0.460    \\ %
   \hline
  gain (\%)   &      &     & \color{OliveGreen}{+21.5} & \color{OliveGreen}{+9.5} & \color{OliveGreen}{+14.6} & \color{OliveGreen}{+12.9} & \color{OliveGreen}{+28.37} & \color{OliveGreen}{+18.8} & \color{OliveGreen}{+15.0}    \\ 

\end{tabular}
}
\caption{Ablation of different versions of our model compared with SatCLIP as the base model. Our soft selection using weighted averages over the database performs better than top-k selection. We also find that a single temperature works well for most tasks. Fine-tuning the temperature per task is generally not required as the gain is marginal. The last row shows the difference between the base model and RANGE$^+$.}
\label{table:ablate}
\end{table*}

\begin{table}
\centering

\begin{tabular}{l|cccc}
\hline
                    &top-1 &top-3 &top-5& top-10   \\ \hline

Img & 66.1 & 83.3 & 88.0 & 92.2   \\ 
\hline
Img+CSP & 72.9 & 87.9 & 91.6  & 94.8 \\
Img+GeoCLIP & 72.9 & 88.2 & 91.9 & \underline{95.2}  \\
Img+CSP\_INat$^*$ & 74.4 & 88.8 & 92.2 & 94.9   \\ 
Img+SatCLIP & 75.1 & 88.7 & 91.9 & 94.5   \\ 
\hline
Img+RANGE & \textbf{75.2} & \textbf{89.6} & \textbf{92.9} & \textbf{95.5}   \\ 
Img+RANGE$^+$ & \underline{75.1} & \underline{89.5} & \underline{92.8} & \textbf{95.5} \\
\end{tabular}
\caption{\textbf{Top-k classification accuracy on INat-2018 test split:} Location information acts as a strong prior in fine-grained species classification. Our method shows competitive top-k results against state-of-the-art models.}
\label{table:inat}
\end{table}
\begin{figure}[t]
\begin{center}
\includegraphics[width=\linewidth, scale=0.3]{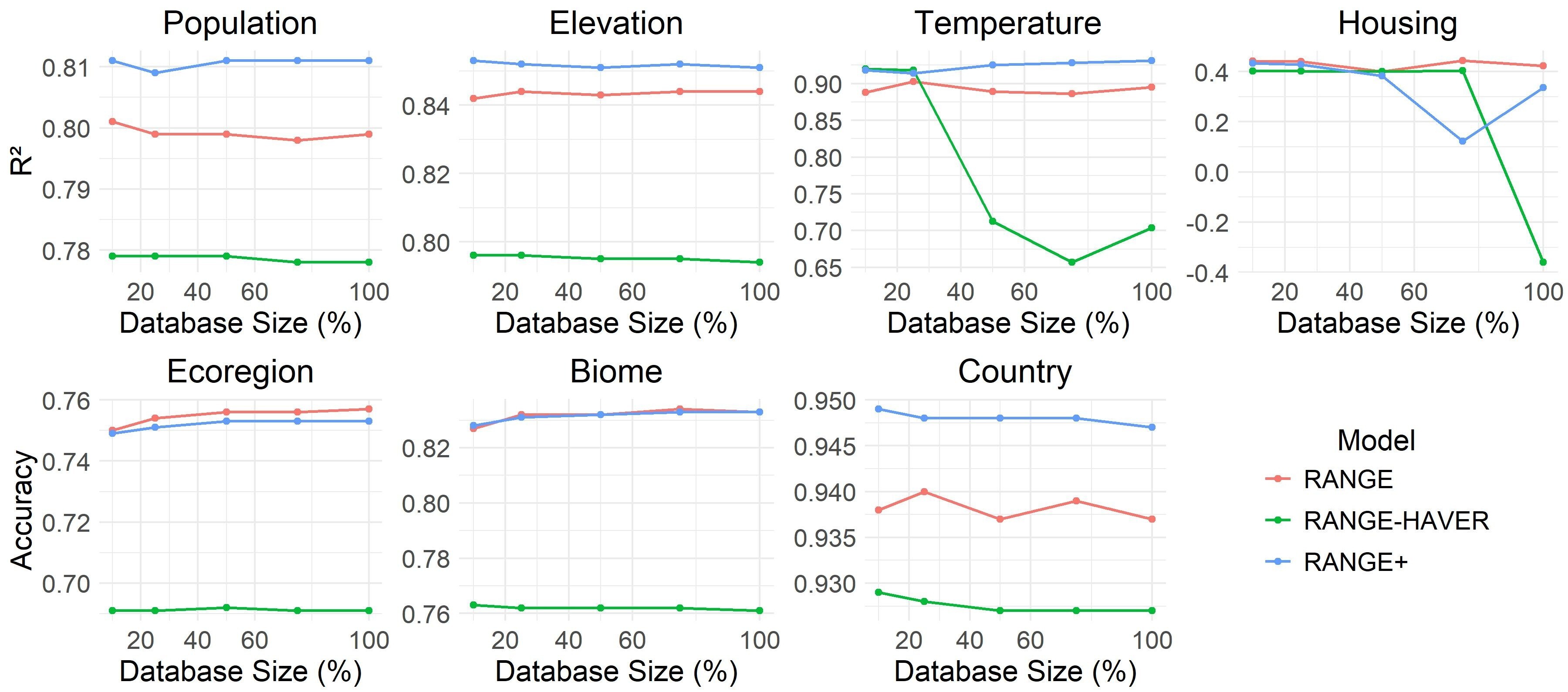}
\end{center}
   \caption{Performance of our model with respect to the database size. The results show that compared to RANGE-HAVER, both RANGE and RANGE$^{+}$ are very robust to changes in database size. We can maintain the same performance even when only using 10\% of the samples in the database.} 
\label{fig:database_ablation}
\end{figure}

\begin{figure*}[!ht]
\begin{center}
\includegraphics[width=\linewidth, scale=0.3]{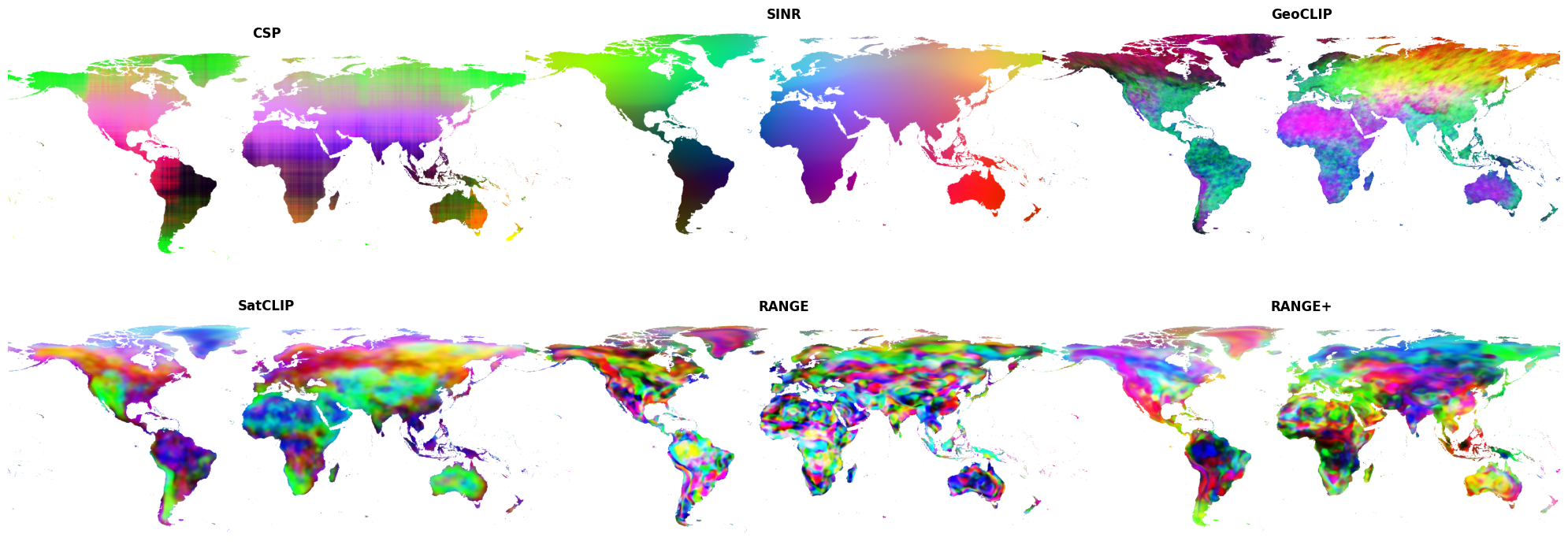}
\end{center}
   \caption{We visualize the geo-embeddings from different models by projecting them into a 3-dimensional vector using Independent Component Analysis (ICA). The results suggest that by explicitly adding visual features, our method learns more high-frequency information compared to the existing models.} 
\label{fig:ica}
\end{figure*}
\begin{figure*}[!t]
\begin{center}
\includegraphics[width=\linewidth, scale=0.3]{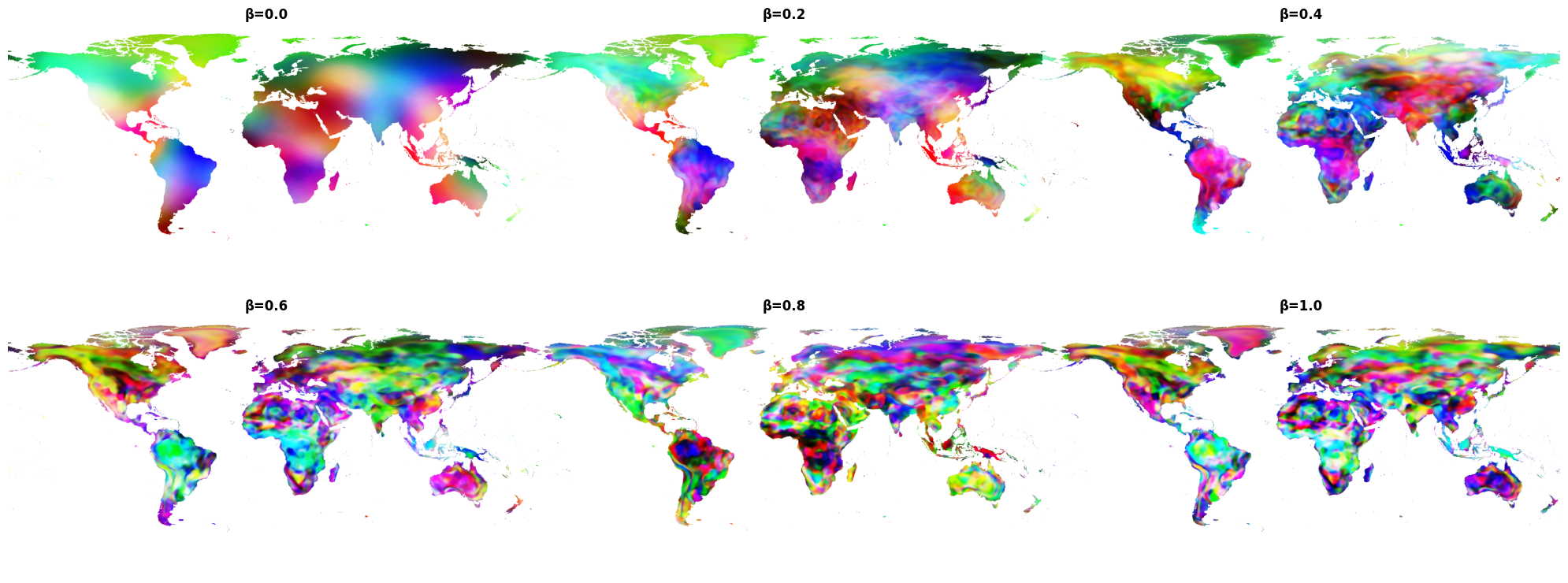}
\end{center}
   \caption{Interpolating the $\beta$ parameter in RANGE$^+$ allows us to control the spatial smoothness of our embeddings. The results show that RANGE$^+$ can be used to generate neural fields of geo-embeddings at multiple frequencies.} 
\label{fig:beta}
\end{figure*}

\section{Experiments and Results}
\subsection{Quantitative Evaluation on Downstream Tasks}
\label{sec:quant}
We evaluate the efficacy of RANGE and RANGE$^+$ embeddings on a wide variety of downstream applications. We create our database using the SatCLIP~\cite{klemmer2023satclip} dataset with 82k locations. For RANGE$^+$, we set $\beta$ to $0.5$. We choose 3 classification tasks and 4 regression tasks for this experiment as used by prior work~\cite{klemmer2023satclip}. The classification tasks are biome classification~\cite{dinerstein2017ecoregion}, ecoregion classification~\cite{dinerstein2017ecoregion}, and country classification~\cite{klemmer2023satclip}. The regression tasks are air temperature prediction~\cite{Hooker_Duveiller_Cescatti_2018}, elevation prediction~\cite{rolf2021generalizable}, population density prediction~\cite{rolf2021generalizable}, and housing price prediction~\cite{pace1997sparse}. Results on climate data from ERA5 are shown in Section~\ref{sec:era5} of the supplementary material. The objective is to learn a linear predictive model that solves the underlying task. For fairness, we use the same hyperparameters for all tasks, i.e., the same $\tau$ and $\beta$ parameters for each task. The details of the hyperparameters are listed in the supplementary material. We compare our method with state-of-the-art parametric~\cite{klemmer2023satclip, vivanco2024geoclip, cole2023spatial, mai2023csp, mai2023sphere2vec} and non-parametric~\cite{Aodha_2019_ICCV,gaolearning} representations. The results of this experiment are shown in Table~\ref{table:main}. 

The results show that our method outperforms all state-of-the-art models in the classification tasks with significant margins. In biome, ecoregion, and country classification, we advance the state-of-the-art by 13.1\%, 4,1\%, and 6.4\%, respectively. Among the regression tasks, we achieve SoTA performance for elevation and population density prediction with significant gains and narrowly come second in the air temperature prediction task. Our method, however, underperforms in the housing price prediction task. We suspect this is because the visual features that define housing change dramatically over time. The standard California housing dataset is based on a 1990 census, while our retrieved visual features are extracted from 2020 Sentinel data. 
The results also demonstrate that the hyperparameters for our model are robust and do not need to be tuned for specific tasks to achieve good performance. Furthermore, RANGE$^+$, on average, outperforms RANGE by a small margin, which can be attributed to the spatial smoothness constraint.

\subsection{Using Location Embeddings as Geo-prior}
\label{sec:inat}
We evaluate our model on the task of fine-grained species classification. We use the iNaturalist 2018 dataset~\cite{van2018inaturalist}, which contains data from 8,142 species. Prior works~\cite{Aodha_2019_ICCV,russwurm2023geographic,klemmer2023satclip} have shown that adding location information can improve the performance of pretrained image classification models on iNat as knowing the location of an image can provide a strong prior about its content. For our experiment, we follow a similar setup as~\cite{russwurm2023geographic}. We take a pretrained image classification model, which gives us $P(y \mid I)$. We then train a linear model that takes as input the pre-trained geo-embeddings and predicts the species categories, i.e., $P(y \mid G)$. The final distribution is computed as the product of the two distributions, i.e., $P(y\mid I, G) = P(y\mid I)*P(y\mid G)$. This experiment also aims to show that location representations obtained using self-supervised training can be easily generalized towards tasks like iNat. Hence, we exclude SINR~\cite{cole2023spatial} from this comparison as it is trained in a supervised manner on the iNaturalist data itself.   

The results of this experiment are shown in Table~\ref{table:inat}.   The performance of the pretrained image classifier is 66.1\%. The addition of location information improves the accuracy across all models. We see that RANGE and RANGE$^+$ achieve competitive scores, narrowly outperforming state-of-the-art models. The results from this experiment highlight the benefits of using good location representations as geo-prior to solve fine-grained image classification tasks.

\subsection{Ablation Study}
\label{sec:ablation}
\textbf{Robustness to Database Size: }The RANGE database that we used for our model in Section~\ref{sec:quant} was created using the SatCLIP dataset with 82k locations. However, our method is extremely robust and can be used with much smaller database sizes. We conducted an experiment where we used smaller samples of the SatCLIP data to create our RANGE database. We created a database with 75\%, 50\%, 25\%, and 10\% of the total data. We then trained a linear model for RANGE and RANGE$^+$ using each of these subsets as our retrieval database. We also trained another version of our method where we only use geodesic distance for retrieval,i.e. $\beta=0$, which we call RANGE-HAVER. We used the same tasks described in Section~\ref{sec:quant} to evaluate these models.

Our results from figure~\ref{fig:database_ablation} show that we can obtain similar performance while using only 10\% of the original database size. In terms of storage, this database only occupies 85 Megabytes in disk, which is very efficient. The results also support our initial claim that there is low variance in satellite images; thus, a small number of satellite images can capture many important semantics across the globe. We see that the performance for both RANGE and RANGE$^+$ is stable across different settings and different tasks. The results also show that RANGE-HAVER, which only relies on spatially retrieved information, can be unstable, as seen in figure~\ref{fig:database_ablation}. These results highlight the pitfalls of depending completely on spatial similarity for retrieval. Locations that are spatially close do not necessarily share similar semantics. Therefore, as you change the sample distribution in the database, you can introduce noisy retrievals, which amplify the noise in the estimated visual embedding. Semantic-based retrieval strategy mitigates such issues.
\\

\noindent \textbf{Impact of architectural components: }Table~\ref{table:ablate} shows the impact of different components of our method compared to SatCLIP as the base model. We also compare our method with two other methods of retrieval: top-1 and top-k (where k was set to 100). The results show that our retriever function generally yields better results than the top-k selection approach. This can be attributed to the fact that top-k averaging does not consider the measure of similarity between query location and keys. We also performed a linear search to find the optimal temperature parameters of RANGE for each task separately. Our results show we are able to make marginal improvements by tuning the temperature parameter for individual tasks. The results indicate the temperature parameters of RANGE are robust and do not need to be finetuned for individual tasks in most cases.

\subsection{Qualitative Evaluation of Geo-embeddings}
\label{sec:ica}
\textbf{Visual comparison of geo-embeddings: }We qualitatively compare the RANGE embeddings with other location embeddings. We use Independent Component Analysis (ICA) to project the embeddings to a 3-dimensional vector and use them as RGB channels to visualize them over the globe. Figure~\ref{fig:ica} shows the visualizations for each model. Qualitatively, we see that models like CSP and SINR are extremely smooth, suggesting that they predominantly capture low-frequency information. GeoCLIP and SatCLIP capture relatively high-frequency information. The visualizations further suggest that RANGE and RANGE$^+$ embeddings are able to capture even higher frequency information. 
\\
\noindent \textbf{Visualizing the impact of $\beta$ parameter: }Secondly, we visualize the impact of the $\beta$ parameter in equation~\eqref{eq:rangep}. A quantitative evaluation of the $\beta$ parameter is presented in Section~\ref{sec:beta_quant} of the supplementary material. Setting $\beta=1$ gives us the RANGE embeddings, setting $\beta=0$ gives us RANGE-HAVER embeddings, and setting $\beta=0.5$ gives the RANGE$^+$ embeddings. This parameter controls the contribution of spatially-retrieved visual information to the semantically-retrieved visual information. This additional information acts as a spatial smoothness constraint on the RANGE embeddings. At $\beta=0$, we add the maximum constraint, which gives us low-frequency embeddings similar to CSP and SINR. As we increase the value of $\beta$, we increase the frequency of our embeddings, i.e. locations that are spatially close to each other but are semantically different have different representations. Thus, the $\beta$ parameter allows us to explicitly control the smoothness of the RANGE embeddings and, therefore, allows generating location embeddings at multiple frequencies.   

%% file: sec/5_conclusion.tex
\section{Discussion and Conclusion}

Self-supervised learning has been a key enabler of the rapid recent progress in computer vision and natural language processing. 
Representations from models like SimCLR~\cite{chen2020simple}, and CLIP~\cite{radford2021learning} have facilitated the creation of many powerful vision and language models. Recent works like SatCLIP~\cite{klemmer2023satclip}, GeoCLIP~\cite{vivanco2024geoclip}, and CSP~\cite{mai2023csp} showed that the same self-supervised techniques can be used to learn powerful representations for geographic locations, which are useful for a broad range of tasks.

In this paper, we introduced a simple, yet effective strategy for improving such geographic embeddings, moving past previous assumptions of multiview redundancy. We proposed a retrieval-augmented strategy for estimating visual features from an auxiliary database of visual embeddings. Our approach results in significant improvements for a variety of tasks as compared to purely parametric embedding strategies with only a modest increase in storage requirements.
Our method is efficient, robust, and multi-scale. We hope our insights and results substantiate our recommendation of using RANGE as a general-purpose location encoder for geospatial tasks. 

\section{Acknowledgments}
This research used the TGI RAILs advanced compute and data resource which is supported by the National Science Foundation (award OAC-2232860) and the Taylor Geospatial Institute.

%% file: sec/X_suppl.tex
\clearpage
\setcounter{page}{1}
\maketitlesupplementary

\section{Implementation Details}
\label{sec:imp}
In this section, we describe the specifics of our experiments. 
\\
\\
\noindent\textbf{Quantitative evaluation on downstream tasks: }For Section~\ref{sec:quant}, we conducted a linear probe for each model using a ridge classifier. We swept over different regularization weights and selected the optimal one using cross-validation on the training set. For our RANGE model, we used $\tau=1/15$ for all tasks. This value was selected by using the cross-validation scores on only Biome and Temperature data. For RANGE$^+$, we used $\tau_1=1/12$, and $\tau_2=1/40$. These were selected using the same procedure as we described for RANGE. For all the experiments with RANGE$^+$, we set $\beta=0.5$, giving equal weight to the semantic and spatial similarity of visual features.
\\
\\
\noindent\textbf{Evaluation on iNaturalist data: }For Section~\ref{sec:inat}, we conducted a linear probe for each model using the training split of iNaturalist data. However, following prior work~\cite{russwurm2023geographic}, we used the ``assume negative" loss function, proposed by Cole \textit{et al}~\cite{cole2023spatial}. For RANGE and RANGE$^+$, we use the same hyperparameters as we used for our experiments in Section~\ref{sec:quant}. We used the pre-trained ``full high-resolution" model by Mac Aodha \textit{et al.}~\cite{Aodha_2019_ICCV} to get the image-only predictions $P(y \mid I)$ for the iNaturalist test set.
\\
\\
\noindent\textbf{Ablation of database size: } For the ablation on database sizes, we used a stratified sampling strategy to create smaller databases with 75\%, 50\%, 25\%, and 10\% of the original data. The original data contained around 82,000 locations uniformly distributed across the landmass. The 82k locations are a subset of the SatCLIP~\cite{klemmer2023satclip} dataset after removing corrupted downloads. We use fixed hyperparameters for the models across all tasks while varying the database size.
\\

\section{Quantitative Evaluation of $\beta$ parameter}
\label{sec:beta_quant}
In this section, we show how the $\beta$ parameter can be tuned to solve geospatial tasks at different resolutions. To show this, we use the checkerboard experiment, which was used by Ru{\ss}wurm \textit{et al.}~\cite{russwurm2023geographic}. We choose k points in the sphere using Fibonacci-lattice; the surface area represented by each point is almost identical~\cite{russwurm2023geographic, gonzalez2010measurement}. Each of these points is assigned one out of 16 categories in a regular order. For the train and test set, we sample 10,000 points on the sphere and assign each point the label of the closest labeled point. The task is to learn a linear model to classify each point (we use the same strategy described in Section~\ref{sec:imp}).

Changing k allows us to change the spatial scale of the task. Higher k creates more grid cells and, therefore, requires higher resolvable resolution. We use different $\beta$ values to solve the checkerboard task with different k's. The results are shown in Table~\ref{table:beta}. The columns in the table shows the different values of k and the average distance between the checkerboard centers in degrees. We see that as we increase the resolution of the task, increasing the value of $\beta$ (reducing spatial smoothness) achieves better performance. Similarly, lower $\beta$ (adding spatial smoothness) performs better for low-resolution tasks. 

We outperform the existing baselines across all resolutions. Within the baselines, SINR performs the best at lower resolutions, whereas SatCLIP performs better at higher resolutions. The quantitative results validate the qualitative results from Section~\ref{sec:ica}. RANGE$^+$ outperforms all the baselines across all spatial resolutions. At $\beta=0.5$, we get the most stable performance across different resolutions. We also see that the gap between the state-of-the-art baseline and RANGE$^+$ increases more dramatically for higher resolutions.

\begin{table}[]
\centering
\begin{tabular}{l|c|ccccc}
\textbf{} & $\beta$ & \begin{tabular}[c]{@{}c@{}}100\\ \footnotesize 19.21$^\circ$\end{tabular} & \begin{tabular}[c]{@{}c@{}}500\\ \footnotesize 8.72$^\circ$\end{tabular} & \begin{tabular}[c]{@{}c@{}}1000\\ \footnotesize 6.11$^\circ$\end{tabular} & \begin{tabular}[c]{@{}c@{}}1500\\ \footnotesize 5.06$^\circ$\end{tabular} & \begin{tabular}[c]{@{}c@{}}2000\\ \footnotesize 4.34$^\circ$\end{tabular} \\ \hline

SatCLIP &  & 36.0 & 26.4 & 25.5 & 25.5 & 21.7 \\
GeoCLIP &  & 64.1 & 22.8 & 15.7 & 14.4 & 13.8 \\
CSP-INat &  & 44.9 & 27.1 & 24.0 & 21.9 & 18.5 \\
CSP &  & 67.8 & 33.8 & 26.8 & 23.9 & 21.1 \\
SINR &  & 87.6 & 58.4 & 34.3 & 22.6 & 19.0 \\
\hline
 & 0       & \textbf{94.0}                                                                            & \textbf{73.2}                                                                           & 50.0                                                                                     & 43.1                                                                                     & 37.8                                                                                     \\
          & 0.25    & \underline{93.3}                                                                               & \underline{72.6}                                                                              & \underline{55.8}                                                                               & 50.7                                                                                     & 45.3                                                                                     \\
   RANGE$^+$       & 0.5     & 92.3                                                                                     & 70.0                                                                                    & \textbf{56.4}                                                                            & \textbf{52.3}                                                                            & \textbf{47.2}                                                                            \\
          & 0.75    & 89.4                                                                                     & 65.1                                                                                    & 54.5                                                                                     & \underline {50.6}                                                                               & \underline {46.3}                                                                               \\
          & 1       & 65.7                                                                                     & 53.6                                                                                    & 50.5                                                                                     & 46.6                                                                                     & 42.6                                                                                    
\end{tabular}
\caption{We quantitatively show that controlling the beta parameter allows us to generate optimal embeddings depending on the resolution of the task. We evaluate on the checkerboard task~\cite{russwurm2023geographic} and change the number of grid cells in the Fibonacci lattice to simulate tasks with different spatial resolutions. Lower $\beta$ yields better embeddings for low-resolution tasks, whereas higher $\beta$ yields better embeddings for high-resolution tasks. We see that we outperform all the baselines across all spatial resolutions.}
\label{table:beta}

\end{table}

\begin{table*}[!ht]
\centering
\begin{tabular}{l|cccccccc|c}
Models    & \multicolumn{1}{l}{temp\_mean} & \multicolumn{1}{l}{temp\_min} & \multicolumn{1}{l}{temp\_max} & \multicolumn{1}{l}{dew\_temp} & \multicolumn{1}{l}{precipitation} & \multicolumn{1}{l}{pressure} & \multicolumn{1}{l}{u\_wind} & \multicolumn{1}{l|}{v\_wind} & \multicolumn{1}{l}{Avg} \\ \hline
CSP       & 0.944                               & 0.933                              & 0.940                              & 0.918                         & 0.610                             & 0.427                                 & 0.499                       & 0.550                        & 0.727                       \\
CSP-INat & 0.987                               & 0.897                              & 0.886                              & 0.857                         & 0.534                             & 0.307                                 & 0.413                       & 0.386                        & 0.658                       \\
SINR      & \underline{ 0.982}                         & \underline{ 0.975}                        & \underline{ 0.976}                        & \underline{ 0.977}                   & 0.758                             & 0.706                                 & 0.726                       & 0.694                        & 0.849                       \\
GeoCLIP   & 0.960                               & 0.953                              & 0.948                              & 0.954                         & 0.591                             & 0.651                                 & 0.502                       & 0.529                        & 0.761                       \\
SatCLIP   & 0.904                               & 0.900                              & 0.887                              & 0.894                         & 0.497                             & 0.743                                 & 0.488                       & 0.455                        & 0.721                       \\ \hline
RANGE     & 0.975                               & 0.972                              & 0.966                              & 0.972                         & \underline{ 0.759}                       & \underline{ 0.888}                           & \underline{ 0.741}                 & \underline{ 0.717}                  & \underline{ 0.873}                 \\
RANGE$^+$    & \textbf{0.990}                      & \textbf{0.985}                     & \textbf{0.984}                     & \textbf{0.988}                & \textbf{0.815}                    & \textbf{0.896}                        & \textbf{0.742}              & \textbf{0.772}               & \textbf{0.896}             
\end{tabular}
\caption{We show the linear probe results on real-world climate data from ERA5. We predict 8 different climate variables using different location encoders. The results show that RANGE and RANGE$^+$ achieve the two highest average R$^2$ across all variables, with RANGE$^+$ consistently achieving the best performance for each task.}
\label{table:era5}
\end{table*}

\begin{figure*}[!ht]
\begin{center}
\includegraphics[width=\linewidth, scale=0.3]{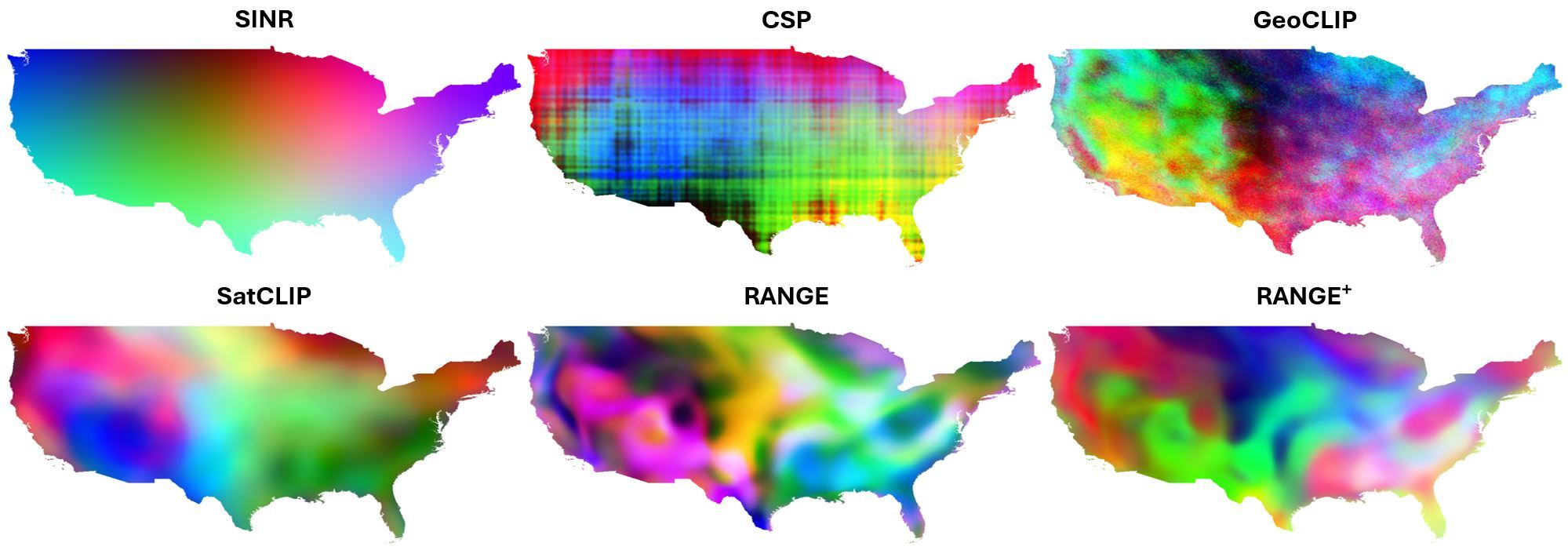}
\end{center}
   \caption{We visualize the geo-embeddings from different models on a country scale (USA) by projecting them into a 3-dimensional vector using Independent Component Analysis (ICA).} 
\label{fig:usa}
\end{figure*}

\section{Evaluation on ERA5 data}
\label{sec:era5}
We also evaluate our models on climate data from ERA5. We use 8 climate variables, namely, mean air temperature, maximum air temperature, minimum air temperature, dewpoint temperature, precipitation, surface pressure, u component of the wind and v component of the wind. We fit a linear model to predict each of these variables using the location embeddings from different location encoders. We use the same hyperparameters for RANGE and RANGE$^+$ that are described in Section~\ref{sec:imp}. Table~\ref{table:era5} shows the R$^2$ values for each task from each model. We show that RANGE and RANGE$^+$ achieve the two highest average R$^2$ across all tasks. Furthermore, RANGE$^+$ also achieves the highest R$^2$ value for each task separately.     

\section{Visualizing Geo-Embeddings at Country Scale}
In Section~\ref{sec:ica}, we visualized the location embeddings on a global scale. Here, we visualize the location embeddings on a country scale. We densely sample points across the United States and use them to compute the location embeddings. We use Independent Component Analysis to project each embedding to a 3-dimensional vector and use it to represent the RGB channels. For different models, the same colors do not necessarily indicate similar information. We can see the visualizations in Figure~\ref{fig:usa}. Visually, it appears that the RANGE embeddings can capture local variations relatively well. 

\section{Geoprior Evaluation with Training-free Baselines}
In Section~\ref{sec:inat}, we evaluated different training-based location encoding methods on geoprior task using iNaturalist data. Here, we show the results of using training-free location encoding methods. Table~\ref{table:training_free} shows that RANGE models outperform the training-free baselines. 

\begin{table}[]
\begin{tabular}{l|llll}
          & top-1 & top-3 & top-5 & top-10 \\ \hline
Direct    & 63.5  & 81.7  & 87.0  & 91.8   \\
Cartesian & 65.0  & 82.7  & 87.7  & 92.3   \\
Wrap      & 65.4  & 83.1  & 87.9  & 92.5   \\
SphereC+  & 69.5  & 85.5  & 89.9  & 93.5   \\
SphereM+  & 70.8  & 86.3  & 90.5  & 93.9   \\ \hline
RANGE     & \textbf{75.2}  & \textbf{89.6}  & \textbf{92.9}  & \textbf{95.5}   \\
RANGE+    & \underline{75.1}  & 
\underline{89.5}  & \underline{92.8}  & \textbf{95.5}  
\end{tabular}
\caption{\textbf{Top-k classification accuracy on INat-2018 test split:} Comparing our model with training free baselines.}
\label{table:training_free}
\end{table}

\section{Computational Cost}
The retrieval process incurs some added computational cost. However, our setup makes this process highly efficient. Generating the feature bank is a one-time operation, which is inexpensive for a few thousand images (Section~\ref{sec:ablation}). Second, the retrieval process is completely vectorized and highly efficient. For reference, when using the 10k database, computing the RANGE embeddings for 1 million input locations takes less than 65 seconds on our CPU and less than 10 seconds on our H100 GPU, making our method efficient for any practical usage.

\section{Limitations and Future Work}
In our work, we argued the limitations of learning geo-embeddings by contrastively aligning location and images from the perspective of multi-view redundancy. While the aforementioned problems exist for any location-image alignment, we propose a solution for improving location and \textbf{\textit{satellite-image}} alignment. In this paper, we exploit specific properties of satellite data to circumvent the existing issues. In the future, we would like to extend this work to all location-image alignment settings. 